\pdfoutput=1

\documentclass[11pt]{article}

\usepackage[final]{acl}

\usepackage{times}
\usepackage{latexsym}

\usepackage[T1]{fontenc}

\usepackage[utf8]{inputenc}

\usepackage{microtype}

\usepackage{inconsolata}
\usepackage{graphicx}
\usepackage{amsmath}
\usepackage{amsfonts,amssymb}
\usepackage{booktabs}
\usepackage{adjustbox}
\usepackage{epstopdf}
\usepackage{colortbl}
\title{MORE: Multi-mOdal REtrieval Augmented Generative Commonsense Reasoning}

\author{Wanqing Cui, Keping Bi\thanks{Corresponding author.}, Jiafeng Guo, Xueqi Cheng \\
        CAS Key Lab of Network Data Science and Technology, \\ 
        Institute of Computing Technology, Chinese Academy of Sciences, Beijing, China\\
        University of Chinese Academy of Sciences, Beijing, China\\
        \texttt{\{cuiwanqing18z, bikeping, guojiafeng, cxq\}@ict.ac.cn} \\}

\begin{document}
\maketitle
\begin{abstract}

Since commonsense information has been recorded significantly less frequently than its existence, language models pre-trained by text generation have difficulty to learn sufficient commonsense knowledge. Several studies have leveraged text retrieval to augment the models' commonsense ability. Unlike text, images capture commonsense information inherently but little effort has been paid to effectively utilize them. 
In this work, we propose a novel \textbf{M}ulti-m\textbf{O}dal \textbf{RE}trieval (MORE) augmentation framework, to leverage both text and images to enhance the commonsense ability of language models. Extensive experiments on the Common-Gen task have demonstrated the efficacy of MORE based on the pre-trained models of both single and multiple modalities. 
\end{abstract}

\section{Introduction}

\begin{figure}
\begin{center}
  \centerline{\includegraphics[width=\linewidth]{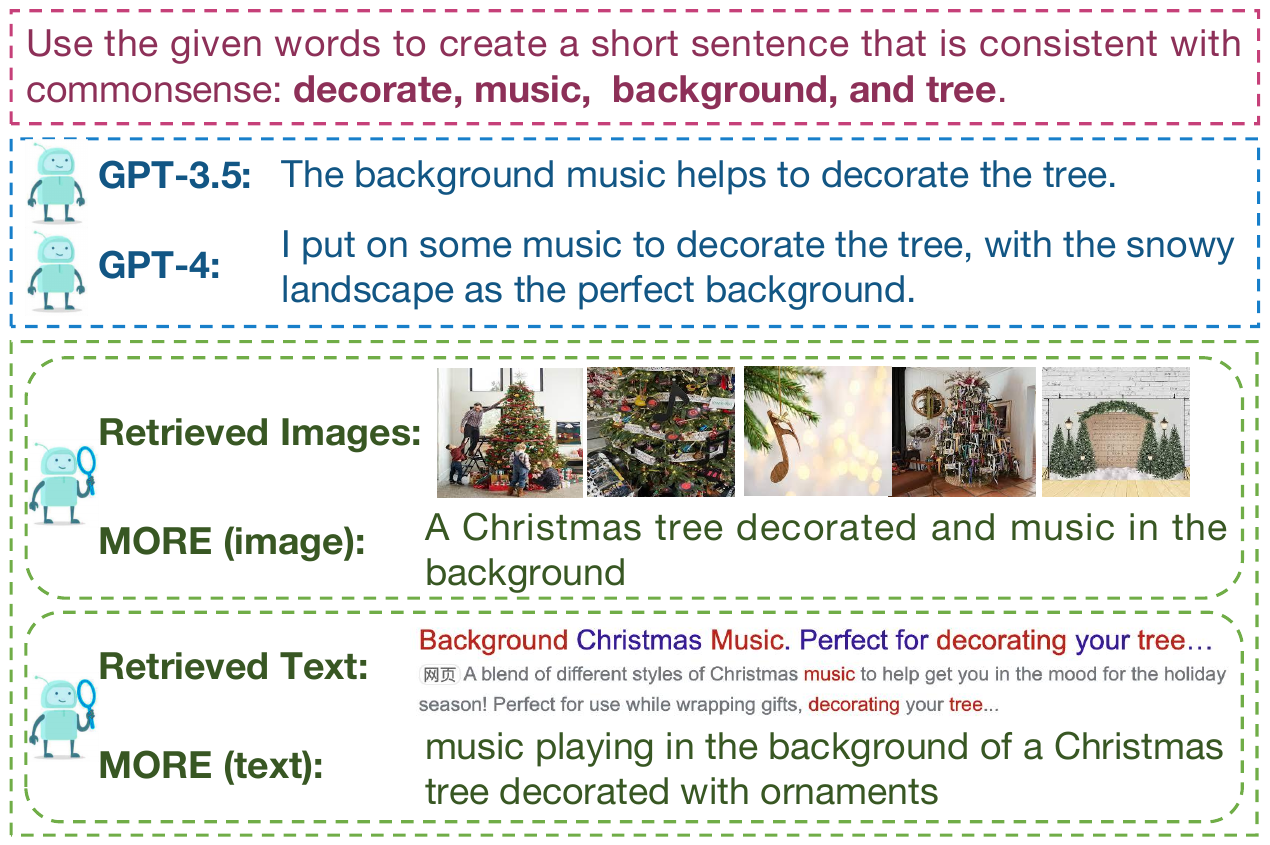}}
  \caption{Sentences made by GPT3.5, GPT-4, and MORE given some concept words.}
\label{fig:intro}
\end{center}
\end{figure}

Language Models (LMs) have gained increasing prominence in artificial intelligence, especially Large Language Models (LLMs) such as LLaMA~\cite{Touvron2023LLaMAOA}, GPT-3.5~\cite{OpenAI2022IntroChat}, and GPT-4~\cite{achiam2023gpt} that have achieved compelling performance across various tasks. However, even LLMs still lack robust commonsense capabilities and can sometimes generate sentences that violate commonsense knowledge. Figure~\ref{fig:intro} illustrates an instance of composing a sentence given several words, where both GPT-3.5 and GPT-4 consider that music can decorate the tree, which makes nonsense. 

Due to the well-recognized reporting bias~\cite{Gordon2013ReportingBA}, i.e., the recording of commonsense information is significantly less than its existence in reality~\cite{grice1975logic, havasi2007conceptnet}, it is inherently difficult for LMs to learn enough commonsense knowledge from modeling text generation. To enhance their commonsense ability, there have been a few attempts to retrieve external commonsense text information~\cite{he2022metric, Li2021KFCNetKF, liu2022kgr4} to augment the LM generation, which have been shown to be effective on commonsense reasoning tasks. 

In contrast to text, commonsense knowledge is naturally recorded in the visual data. Additionally, text is used primarily for communication and may include subjective statements while images often record the physical world more objectively. Thus, images can be supplementary to text for LMs to enhance commonsense abilities. This can also be confirmed by the fact that humans acquire knowledge from both textual and visual data~\cite{Gambrell1986MentalIA, Bloom2000HowCL, joffe2007comprehension}. Aware of this, instead of retrieving text snippets to assist the models in conducting commonsense tasks ~\cite{he2022metric, yu2022retrieval, Li2021KFCNetKF, liu2022kgr4}, we propose a \textbf{M}ulti-m\textbf{O}dal \textbf{RE}trieval (MORE) augmentation framework to incorporate both text and images. For LLMs pre-trained with multi-modal data (e.g., BLIP2~\cite{Li2023BLIP2BL}), multi-modal retrieval augmentation can also be beneficial since it explicitly provides the text snippets and images carrying related commonsense information to the current sample. To effectively incorporate the multi-modal information into LMs, there are two major challenges:

\textbf{1) How can we enable LMs to effectively extract useful knowledge from multi-modal retrieved results?} This is even more challenging for text-based LMs because of the modality differences. To address this challenge, we propose a plug-and-play integrator that adopts a cross-attention mechanism to weigh each of the multi-modal results based on the query input and extract beneficial information. For text-based LMs, we employ a multi-modal encoder (e.g., the Qformer of BLIP2) to ingest results of images and text. In this case, the integrator also acts as a bridge that transforms the encoded semantic space of the retrieved results into the representation space used by the LMs.

\textbf{2) Since the retrieval quality could vary considerably, how can we ensure the LMs do not ignore the retrieved results and also not trust them blindly?} On the one hand, to prevent LMs from disregarding the entire retrieved results due to the noise they may contain, we introduce a training mechanism in MORE, i.e., query dropout, that masks the query input to the LMs at a certain ratio to urge the LMs to leverage the retrieved results for generation. On the other hand, to avoid too much dependence on the results that could be noisy, when queries are dropped out, we randomly replace the results with irrelevant ones and guide the LMs to output empty in such cases, so that the LMs can learn that it is not necessary to use retrieval all the time. 

We evaluate our approach on a generative commonsense reasoning task, i.e., CommonGen~\cite{lin2020commongen}. This task requires models to generate reasonable sentences using given concepts. 
Experimental results show that MORE can significantly boost the performance on CommonGen by incorporating multi-modal retrieved results for the LMs pre-trained with data of single or multiple modalities. MORE also significantly outperforms representative retrieval augmentation baselines and LLMs like GPT-3.5 and GPT-4, demonstrating the effectiveness of its architecture and training mechanism.

We summarize our contributions as follows: 
(1) We propose a novel multi-modal retrieval augmented language modeling framework for enhancing text generation of LMs.
(2) Evaluations on the generative commonsense reasoning task, i.e., CommonGen, demonstrate the effectiveness of MORE on single/multi-modal LMs.
(3) We conduct comprehensive analyses to verify the effectiveness of MORE under various settings and illustrate its advantages compared to LLMs like GPT-3.5 and GPT-4 through case studies.


\section{Related Work}
\subsection{Retrieval Augmented Generation}
The effectiveness of introducing additional contexts in the generation task has been demonstrated. Specifically, utilizing the input as a query, a retriever initially retrieves a set of documents from a corpus. Then a LM integrates these retrieved documents as supplementary information to generate a final prediction. For instance, Atlas~\cite{izacard2022few} finetunes a LM jointly with the retriever with very few training examples. RETRO~\cite{borgeaud2022improving} modifies the decoder-only architecture to incorporate retrieved texts and pretrains the LM from scratch. Both methods necessitate updating model parameters through gradient descent, a process not applicable to Large Language Models (LLMs). 

Given that the cost of fine-tuning LMs may not always be acceptable, recent research has explored retrieval augmentation for frozen LMs. ~\citet{mallen-etal-2023-trust, si2022prompting, Ram2023InContextRL} demonstrate that directly prepending the documents returned by a frozen retriever to the input can improve LMs performance on open-domain QA. To support a large number of documents, FiD~\cite{izacard2021leveraging} processes each input passage in parallel in the encoder. RePlug~\cite{shi2023replug} further finetunes the retriever based on feedback from the frozen LM to get more helpful retrieved results. On these bases, compressing the retrieved results at the sentence level~\cite{xu2023recomp} or token level~\cite{liu2023tcra, berchansky2023optimizing} can boost performance by filtering irrelevant information retrieved and improve computing efficiency.

\subsection{Image Enhanced Text Generation}

VisCTG~\cite{feng2022retrieve} enhances the commonsense ability of LMs by retrieving images and using image captions as input augmentation. In addition to explicitly retrieving images, VAWI~\cite{Guo2022VisuallyaugmentedPL} leverages information from vision-language models, i.e. CLIP~\cite{radford2021learning}, to aid natural language understanding. I\&V~\cite{wangcontextualized} train an imagination model to generate a scene graph given an input under the supervision of images and then train LMs to generate sentences based on both input and scene graph. The above methods either do not directly use images as non-verbal data or require fine-tuning the whole pre-trained LMs to adapt to visual input. Drawing on the importance of imagination to human writing, iNLG~\cite{zhu2022visualize} and LIVE~\cite{Tang2023LearningTI} use images generated by an image-generative model based on text inputs as supplementary information and train the LM to generate under visual guidance. However, the generated images may not necessarily carry commonsense information, such as cartoon images.

\section{Generative Commonsense Reasoning}

We focus on the task of CommonGen~\cite{lin2020commongen} to investigate and enhance the common sense reasoning capabilities of LMs. 

\begin{figure*}
\begin{center}
  \centerline{\includegraphics[width=0.9\linewidth]{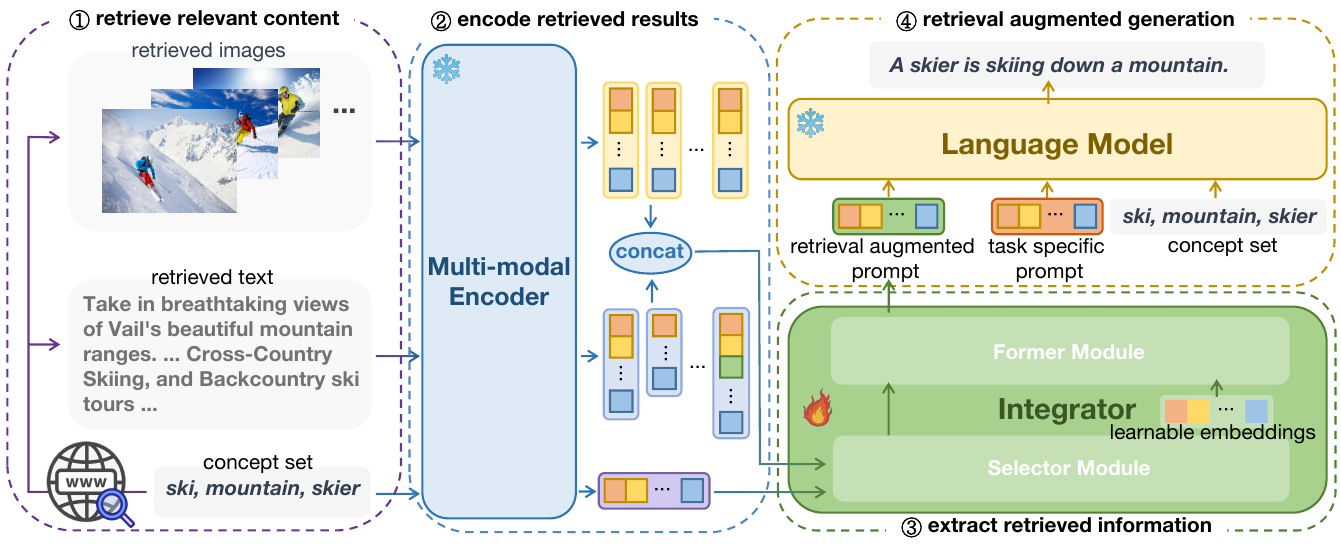}}
  \caption{The process of our framework generating the sentence given input concepts based on multi-modal retrieval augmentation.}
\label{fig:model}
\end{center}
\end{figure*}

\subsection{Preliminaries}
\textbf{Problem Statement:} The generative commonsense reasoning task in CommonGen asks the LM to make a sentence $y$ that contains all the concept words in the given set $C=\{c_1, ..., c_K\}$, where $c_i$ denotes the $i$-th concept and $y$ should describe a common scenario in our daily life. 

\textbf{Training Objectives:} It is usually modeled as a sequence generation task and is optimized by minimizing the cross-entropy loss between the predicted token distribution and the reference distribution: $L = - \sum_{t=1}^{|y|} \log P(y_t|C, y_{<t})$. In this work, to ensure parameter efficiency and applicability to LLMs, we use prompt tuning~\cite{Lester2021ThePO, Liu2021GPTUT} instead of fine-tuning LMs. We only tune a task prompt, which is prepended to the input word embeddings in the first layer.
When retrieval augmentation is enabled, a set of retrieved items $D$, which is retrieved with the concepts as query words, is also used as input and the new loss function becomes:
\begin{equation}
  \begin{aligned}
    L & = - \sum_{t=1}^{|y|} \log P(y_t|C, D, y_{<t}).
  \end{aligned}
\end{equation}

\subsection{Multi-Modal Retrieval Augmention}


As shown in Figure~\ref{fig:model}, the \textbf{M}ulti-m\textbf{O}dal \textbf{RE}trieval (MORE) augmented framework for text generation has four core components: retrieving relevant images and texts based on the concept words (§~\ref{retrieve}), encoding the retrieved results with an encoder (§~\ref{encode}), extracting useful information to yield a retrieval augmented prompt with an integrator (§~\ref{integrate}), and generating sentences based on the retrieval augmented prompt, task prompt, and concept embeddings with the frozen LM backbone (§~\ref{generate}).

\subsubsection{Retrieval Results for Augmentation}
\label{retrieve}

Previous work~\cite{he2022metric, Li2021KFCNetKF, liu2022kgr4} that incorporates retrieval augmentation on this task consider the image/video captions~\cite{krishna2017dense, Williams2017ABC, Wang2019VaTeXAL, Bowman2015ALA, Lin2014MicrosoftCC} that CommonGen is built on as the retrieval candidates, which is obviously impractical. In such a setting, we find that the captions retrieved by BM25~\cite{robertson2009probabilistic} can achieve comparable performance with the state-of-the-art (SOTA) methods that train retrievers for augmenting LLMs (shown in Appendix~\ref{app:bias_result}), making the investigation less meaningful.

To accommodate the task in real-world scenarios, in this paper, we retrieve the image and text results from a general Web search engine, i.e., Bing, for retrieval augmentation. We employ Bing as a reasonable off-the-shelf retriever since our focus is on how to incorporate the supporting items rather than training a powerful retriever. Formally speaking, given a concept set $C$, we retrieved $M$ images and $N$ text snippets by Bing using words in $C$ as the query, comprising the set of items $D=\{d^v_1, ...d^v_M, d^t_1,...,d^t_N\}$. 

Specifically, after we preprocessed the retrieved results by removing duplicate images and noisy text, we collected a total of 500,100 images and 787,970 text snippets. On average, each concept set has 14 images and 23 passages, which means that $M$ and $N$ can be at most 14 and 23 respectively. We retain the order returned by the browser without any re-ranking. See the Appendix~\ref{app:crawl} for more details.

\subsubsection{Multi-Modal Encoder}
\label{encode}

Then we use an encoder to get the initial representation $e_i^{ra}$ for each retrieved image or passage $d_i$:
\begin{equation}
  \begin{aligned}
    e_i^{ra} & = \textrm{Encode}(d_i).
  \end{aligned}
\end{equation}
This results in a sequence of representations with $d_{enc}$ dimension. To align the encodings of text and images in the same semantic space, we adopt a multi-modal encoder - the frozen Q-Former of the pre-trained BLIP2~\cite{Li2023BLIP2BL}. Unlike the other commonly used multi-modal encoder - CLIP~\cite{radford2021learning}, that encodes the input to a single final embedding, the Q-Former of BLIP2 outputs a sequence of embeddings and thus can retain more information. 

\subsubsection{Retrieved Information Integrator}
\label{integrate}

The integrator extracts useful information from the representations of multiple retrieved results and condenses it into a retrieval augmented prompt. The Integrator has a Selector module and a Former module.

\textbf{Selector:} This module extracts useful information from the retrieval representations based on the input concept and outputs a variable-length retrieval augmentation representation. It receives the concatenation of multiple initial representations $e^{ra} = [e^{ra}_1; ...;e^{ra}_{M+N}] \in \mathbb{R}^{(M+N) \times d_{enc}} $ and the embeddings of the concept words $e^c \in \mathbb{R}^{l_{c} \times d_{enc}}$ as input, in which $l_c$ is the number of tokens in the concept set. The Selector is composed of two stacks of identical layers. Each layer consists of a self-attention network, which is used for interaction between retrieved content, a cross-attention network, which is used for interaction between retrieved content and concepts, and a fully connected feed-forward network:
\begin{equation}\label{eqn-2}
  \begin{aligned}
    h^{self}_i & = \textrm{Attn}(h_{i-1} W^Q_i; h_{i-1} W^K_i; h_{i-1} W^V_i)\\
    h^{cross}_i & = \textrm{Attn}(h^{self}_i M^Q_i; E^{ra} M^K_i; E^{ra} M^V_i)\\
    h_i & = h^{cross}_i F_i.
  \end{aligned}
\end{equation}
$\textrm{Attn}(Q, K, V)$ is the multi-head attention layer as in Transformer~\cite{vaswani2017attention}. $W \in \mathbb{R}^{d_{enc} \times d_{int}}$, $M \in \mathbb{R}^{d_{enc} \times d_{int}}$ ,and $ F \in \mathbb{R}^{d_{int} \times d_{enc}}$ are projection matrices, in which $d_{int}$ is the dimension of the hidden states of Integrator, and $h_0$ is set to $e^c$. The output of the Selector module is a variable-length retrieval augmentation representation $h_2 \in \mathbb{R}^{l_c \times d_{int}}$.

\textbf{Former:} This module converts the representation produced by the Selector into fixed-length and projects it into the input embedding space of the LM. This results in the final retrieval augmentation prompt $p^{ra}$. The Former comprises a cross-attention network and a fully connected feed-forward network:
\begin{equation}\label{eqn-3}
  \begin{aligned}
    p^{ra'} & = \textrm{Attn}(qM^{Q'} ; h_2M^{K'} ; h_2M^{V'} )\\
    p^{ra} & = p^{ra'} O,
  \end{aligned}
\end{equation}
in which $q \in \mathbb{R}^{l_q \times d_{int}}$ is a learnable embeddings with fixed-length $l_q$. $O \in \mathbb{R}^{d_{int} \times d_{lm}}$ is the projection matrix used for spatial mapping and $d_{lm}$ is the dimension of the input embedding of the LM.


\subsubsection{Soft Prompt Based Text Generation}
\label{generate}

To ensure training efficiency especially based on LMs, we freeze the parameters of the LMs and adopt the Prompt-tuning~\cite{Lester2021ThePO} technique, which incorporates the fixed-length embeddings produced by the Integrator as a plug-and-play soft prompt. 

During text generation, the LM receives the concatenation of the task-specific prompt $p^{task}$ and the concept set $C$ as input, and generates sentence $y$ as output, denoted as $y = \textrm{LM}([p^{task}; C])$. When using retrieval augmentation, besides the task-specific prompt, we also prepend the retrieval-augmented (RA) prompt to the input. Consequently, the input to the model becomes $[p^{ra}; p^{task}; C]$.

\subsubsection{Training Strategy}
\label{training}

\textbf{Query Concept Dropout}: The retrieval quality can vary considerably, so the model may simply ignore the retrieval input instead of learning to extract useful information. To enhance the utilization of retrieval augmented inputs, we propose a query dropout training strategy. Specifically, we randomly mask the query concept $C$ input to the LMs with probability $p$ in the initial $T$ training steps, and let the model generate sentences only based on retrieved results. Please note that query dropout is only applied to the input of LMs, and $C$ is always input to the Integrator to guide the model in extracting beneficial information from the retrieved results. The probability $p$ decreases as the number of training steps increases: $p = 0.5 \times (1 - \sin(\pi (\min(\frac{t}{T}, 1)-0.5)))$.


\textbf{Noisy RA Input:} It is also important to ensure that the model can learn to ignore noise rather than blindly trust the retrieved results. Therefore, we artificially introduce noise during query dropout by replacing the retrieval input with irrelevant results from other samples and correspondingly changing the target output with an `EOS' token with probability $\hat{p}$. 

\section{Experiments Settings}

\subsection{Dataset}
We validate our method on the CommonGen dataset\footnote{https://inklab.usc.edu/CommonGen/. Under MIT license.}~\cite{lin2020commongen}. It is designed for generative commonsense reasoning tasks involving the composition of discrete concepts into sentences depicting everyday scenarios. The dataset comprises 32,651, 993, and 1,497 unique concept sets for training, development, and testing, respectively. Each concept set has multiple associated gold target sentences, yielding 67,389, 4,018, and 6,042 sentences for reference in total. 
When retrieval augmentation is enabled, we used the retrieved results from Bing as introduced in Section~\ref{retrieve}. We will release the crawled images and text to encourage future research in multi-modal retrieval augmentation under a practical setting for CommonGen. 

\subsection{Methods for Comparisons}

\textbf{Text-based/Multi-modal LMs:} For text-based LMs, we employ \textbf{T5$_{\textrm{BASE}}$} as well as \textbf{T5$_{\textrm{LARGE}}$}~\cite{Raffel2019ExploringTL} to represent small pre-trained LMs, and \textbf{OPT$_{2.7b}$}~\cite{Zhang2022OPTOP} to represent the LLMs.  We also query the close source model \textbf{gpt-3.5-turbo-1106}~\cite{OpenAI2023Gpt35} through API with the prompt "Use the given words to make a short sentence that is consistent with commonsense. Words: \{...\}". For Multi-modal LMs (MLMs) we compare with \textbf{BLIP2-OPT$_{2.7b}$}~\cite{Li2023BLIP2BL}, an open source model, and \textbf{gpt-4-1106-vision-preview}~\cite{OpenAI2023Gpt4}, a close source model. Since MLMs can accept images and text as input, we directly input the retrieved items into MLMs. All the above open source models are based on huggingface\footnote{https://github.com/huggingface} and are under Apache License 2.0.

\textbf{Retrieval Augmented Generation Baselines:} We consider two types of textual retrieval augmented models. One is \textbf{Prepend}~\cite{mallen-etal-2023-trust, si2022prompting, Ram2023InContextRL}, which prepends the top-k text results to the concepts as input. The other one is \textbf{FiD} ~\cite{izacard2021leveraging}, which concatenates each retrieved passage with the concept words separately to encode in parallel for better handling of long text. For the visual retrieval augmented model, we compare with \textbf{VisCTG}~\cite{feng2022retrieve}, which generates a caption for each image with an image captioning model~\cite{Luo2018DiscriminabilityOF} and prepends the captions to the input for augmentation. All the above models use T5$_{\textrm{BASE}}$ as the backbone and are tuned with prompt-tuning.

\textbf{MORE with Various Backbones:} We test MORE with various backbones to explore whether it can be effectively used in different model architectures. Specifically, \textbf{T5$_{\textrm{BASE}}$} and \textbf{T5$_{\textrm{LARGE}}$} represent small LMs and are encoder-decoder architecture. \textbf{BLIP2-OPT$_{2.7b}$} represent MLMs. It should be noted that BLIP2-OPT$_{2.7b}$ is equivalent to OPT$_{2.7b}$ when not receiving image input. Therefore it can also be regarded as a variant based on \textbf{OPT$_{2.7b}$}, which represents LLMs and is decoder-only architecture.

\begin{table*}
\setlength{\belowcaptionskip}{-0.2cm}
\caption{Test results on CommonGen(V1.0). The best results are bolded, and `*' indicates that the results are significantly improved ($p<0.05$) compared to the best baseline model (be underlined) under the significance test. In the last block, we also mark results with ${\dagger}$ that are significantly improved compared with the sub-optimal baseline.}
\label{tab:overall_result}
\begin{center}
\begin{tabular}{lccccc}
    \toprule
    Model & Bleu$_4$ & METEOR & ROUGE$_L$ & CIDEr & SPICE \\
    \midrule
   GPT-3.5 (0-shot) & 21.44 & 28.93 & 48.80 & 13.03 & 26.69\\
   GPT-3.5 (3-shot) & 28.91 & 31.14 & 53.25 & 15.92 & 28.89 \\
   \midrule
   T5$_{\textrm{BASE}}$ & \underline{28.93} & \underline{29.48} & \underline{54.25} & \underline{15.36} & \underline{30.95} \\
   Prepend$_{\textrm{BASE}}$ & 26.68 & 28.43 & 53.29 & 14.39 & 29.95 \\
   FiD$_{\textrm{BASE}}$ & 28.32 & 28.72 & 54.07 & 14.78 & 30.25 \\
   VisCTG$_{\textrm{BASE}}$ & 27.67 & 28.82 & 53.71 & 14.77 & 30.24 \\
   \rowcolor{gray!20} MORE$_{\textrm{T5}_{\textrm{BASE}}}$ (text) & 29.87* & 30.15* & \textbf{55.20}* & 15.79* & 31.57* \\
   \rowcolor{gray!20} MORE$_{\textrm{T5}_{\textrm{BASE}}}$ (image) & 29.98* & 30.21* & 55.07* & 15.92* & 31.63* \\
   \rowcolor{gray!20} MORE$_{\textrm{T5}_{\textrm{BASE}}}$ (multi-modal) & \textbf{30.27}* & 30.28* & 55.18* & \textbf{16.02}* & \textbf{31.94}* \\

   \midrule
   T5$_{\textrm{LARGE}}$ & \underline{31.16} & \underline{30.48} & \underline{55.68} & \underline{16.14} & \underline{31.62} \\
   \rowcolor{gray!20} MORE$_{\textrm{T5}_{\textrm{LARGE}}}$ (text) & 32.03* & 31.05* & 56.14 & 16.37 & 32.00 \\
   \rowcolor{gray!20} MORE$_{\textrm{T5}_{\textrm{LARGE}}}$ (image) & \textbf{32.37}* & \textbf{31.31}* & 56.60* & \textbf{16.67}* & 32.31* \\
   \rowcolor{gray!20} MORE$_{\textrm{T5}_{\textrm{LARGE}}}$ (multi-modal) & 32.29* & 30.90* & \textbf{56.62}* & 16.63* & \textbf{32.34}* \\
   \midrule
   OPT$_{\textrm{2.7b}}$ & 31.53 & 31.43 & 55.95 & \underline{16.76} & 32.24 \\
   BLIP2$_{\textrm{opt-2.7b}}$ (multi-modal) & \underline{31.92} & \underline{31.70} & \underline{56.22} & 16.73 & \underline{32.44} \\
   \rowcolor{gray!20} MORE$_{\textrm{OPT}_{\textrm{2.7b}}}$ (multi-modal) & \textbf{32.78}*$^{\dagger}$ & \textbf{32.15}$^{\dagger}$ & \textbf{57.07}*$^{\dagger}$ & \textbf{17.03}*$^{\dagger}$ & \textbf{32.94}*$^{\dagger}$ \\
    \bottomrule
\end{tabular}
\end{center}
\end{table*}

\subsection{Evaluation Metrics}
To assess the generation performance, we use standard metrics: BLEU~\cite{Papineni2002BleuAM} quantifying the overlap between predictions and references based on n-gram precision and ROUGE~\cite{Lin2004ROUGEAP} measuring the n-gram recall. METEOR~\cite{Banerjee2005METEORAA} is an improved version of BLEU and considers both exact word matches and semantic similarities. CIDEr~\cite{Vedantam2014CIDErCI} focuses on capturing sentence semantic similarity.
SPICE~\cite{Anderson2016SPICESP} quantifies the semantic propositional content of generations by leveraging scene graphs. Please notice that SPICE aligns closely with human evaluation and should be treated as the primary metric. We also incorporate sentence similarity metrics (Sent-Sim) with SimCSE~\cite{gao2021simcse} to measure semantic similarity. We use the entire test dataset to obtain the main results and randomly sample 500 data in the test set to compare with GPT-4 for the sake of a limited budget.

\subsection{Implementation Details}

The same set of hyper-parameters is used for all the models\footnote{Code and data are publicly available at https://github. com/VickiCui/MORE}. We use the AdamW~\cite{Loshchilov2017DecoupledWD} optimizer with $\beta1=0.9$, $\beta2=0.999$ and the weight decay is 0.05. The batch size is selected from \{64, 128\}. Models were trained with at most 20,000 steps with a $1\%$ warm-up period. For retrieval augmentation, we train the model with an additional $T=2000$ steps with query dropout, and the noisy RA input ratio $\hat{p}$ is set to $0.3$. The learning rates of the task prompt and the retrieval augmented prompt are selected from $\{1e-4, 5e-4, 1e-3\}$ and $\{1e-5, 3e-5\}$ respectively. During decoding, we use beam search with size 5. We train the models under each setting with 3 random seeds and choose the best ones according to the performance on the development set for testing. The prompt length of task and retrieval augmentation are both set to 32.

\begin{table*}
\setlength{\belowcaptionskip}{-0.2cm}
\caption{Sentence similarity results on the test data. As there are multiple references for one input, 'Avg' represents the average similarity between the model output and all references, while 'Max' represents the similarity between the model output and the closest reference. The best results are bolded.}
\label{tab:sent-sim}
\begin{center}
\begin{adjustbox}{max width=1.\linewidth}
\begin{tabular}{l|cc|cc|ccc|c}
    \toprule
     & T5$_{\textrm{BASE}}$ & MORE$_{\textrm{T5}_{\textrm{BASE}}}$ & T5$_{\textrm{LARGE}}$ & MORE$_{\textrm{T5}_{\textrm{LARGE}}}$ & OPT$_{\textrm{2.7b}}$ & BLIP2$_{\textrm{opt-2.7b}}$ & MORE$_{\textrm{OPT}_{\textrm{2.7b}}}$ &  GPT-3.5 (3-shot) \\
     \midrule
     Avg & 71.51 & 71.59 & 72.12 & 72.23 & 71.66 & 71.83 & \textbf{72.53} & 70.00 \\
     Max & 82.32 & 83.31 & 83.9 & 84.05 & 83.15 & 83.8 & \textbf{84.15} & 80.08 \\
    \bottomrule
\end{tabular}
\end{adjustbox}
\end{center}
\end{table*}



\section{Results and Analysis}
\subsection{Overall Results}

As shown in Table~\ref{tab:overall_result} and Table~\ref{tab:sent-sim}, after incorporating text and images to LMs, our method can boost the generation performance significantly based on various backbones. Comparing images and text, we find that images are better in improving commonsense ability, and incorporating both of them yields even better performance.

As shown in Table~\ref{tab:500_result}, Although based on a smaller model, MORE can achieve better results than GPT-3.5 and GPT-4. This fully illustrates the effectiveness of our method. Considering that GPT-4 is a multi-modal language model, we also test its performance when retrieval augmented items are provided. However, GPT-4 cannot effectively utilize the retrieved inputs, leading to deteriorated performance. Retrieval augmentation methods for the GPT-4 model are worth exploring in the future. We also tested the model with a specified length limit to avoid the tendency of LLMs to generate longer sentences. The length of each sentence is the average length of the golden references, so the results can be regarded as an upper bound. According to the most critical SPICE metric, length constraints do not lead to better results. Further analysis revealed that length constraints result in the concept coverage decrease, indicating that LLMs face challenges in organizing concepts with simple short sentences.

In terms of incorporating retrieved results, MORE is better than Prepend and FiD, which are textual augmented models. Although previous work found that using captions can have better results, this also risks leaking the answer. The method of directly inputting retrieved results becomes invalid after the retrieval content changes. MORE is also better than VisCTG, which is a visual augmented model. As shown in Appendix~\ref{app:captions}, the generated captions are not always accurate and may lack the required information, so pre-converting images to captions is not a proper approach to leverage images. 

\begin{table}
\caption{Test results compared with closed source LLMs on 500 randomly sampled data. `*' and `$\dagger$' indicate the results are significantly improved compared to GPT-4 (3 shot) and GPT-3.5 (3 shot), respectively. The `$n$i$n$t' means use $n$ images and $n$ text as augmentation. The `lc' means the generation length is explicitly constrained to the average length of the references.}
\label{tab:500_result}
\begin{center}
\begin{adjustbox}{max width=1.\linewidth}
\begin{tabular}{lccc}
    \toprule
    Model & Bleu$_4$ & CIDEr & SPICE\\
    \midrule
    GPT-4 (0-shot) & 28.53 & 16.52 & 30.53\\
    GPT-4 (0-shot \& lc) & 27.87 & 16.89 & 29.11 \\
    GPT-4 (3-shot) & 30.00 & 16.41 & 29.05\\
    GPT-4 (0-shot \& 1i1t) & 19.86 & 12.43 & 26.20 \\
    GPT-3.5 (0-shot) & 22.97 & 13.93 & 27.25 \\
    GPT-3.5 (0-shot \& lc) & 25.54 & 15.62 & 26.75 \\
    GPT-3.5 (3-shot) & 28.35 & 16.14 & 29.13 \\
    \rowcolor{gray!20} MORE$_{\textrm{OPT-2.7b}}$ (1i1t) & 31.81*$^\dagger$ & 17.08*$^\dagger$ & 31.81*$^\dagger$ \\
    \rowcolor{gray!20} MORE$_{\textrm{OPT-2.7b}}$ (3i3t) & \textbf{32.53}*$^\dagger$ & \textbf{17.30}*$^\dagger$ & \textbf{32.81}*$^\dagger$ \\
    \bottomrule
\end{tabular}
\end{adjustbox}
\end{center}
\end{table}

\begin{table}
\caption{Ablation study results based on MORE$_{\textrm{BASE}}$.}
\label{tab:ablation_study}
\begin{center}
\begin{tabular}{lcccc}
    \toprule
    Model & Bleu$_4$ & CIDEr & SPICE\\
    \midrule
    T5 & 28.93  & 15.36 & 30.95 \\
    MORE & 30.27 & 16.02 & 31.94 \\
    \quad w/o concept-input & 29.45 & 15.33 & 31.18 \\
    \quad w/o query-dropout & 29.81 & 15.52 & 31.14 \\
    \quad w/o noisy-RA & 29.54 & 15.42 & 30.88 \\
    \bottomrule
\end{tabular}
\end{center}
\end{table}

\subsection{Ablation Study}

We examine the effectiveness of various components within MORE by creating several variants, selectively removing or substituting each component, with results detailed in Table~\ref{tab:ablation_study}.

First, we replace the concepts that are input into the integrator with a randomly initialized learnable token sequence (w/o concept-input). The drop in performance highlights the importance of using concept words for references when extracting beneficial information from the retrieved results. Second, we remove the query dropout strategy (w/o query-dropout). The performance drop demonstrates its importance in effectively leveraging the retrieved results. Finally, we further exclude the noisy retrieval augmented input (w/o noisy-RA). Performance degradation indicates that blindly trust in retrieved input can harm model performance. It is necessary to explicitly instruct the model to learn to ignore the irrelevant augmentation results.

Please note that when the query dropout strategy is removed, it means that noisy retrieval augmented input is also not contained. However, the results of `w/o query-dropout' is better than the results of `w/o noisy-RA'. This emphasizes the disadvantages of blindly trusting retrieval items, and it is necessary for the model to learn to distinguish irrelevant retrieval input.

\subsection{Analysis}

\begin{table}
\caption{Analyze the influence of additional parameters on the results. All models are based on T5$_{\textrm{BASE}}$. `$n$ pl' means using a prompt with $n$ length. `rand-RA' means training models with irrelevant random search results.}
\label{tab:parameter_analysis}
\begin{center}
\begin{tabular}{lccc}
    \toprule
    Model & Bleu$_4$ & CIDEr & SPICE\\
    \midrule
   T5 (32 pl) & 28.93  & 15.36 & 30.95 \\
   T5 (64 pl) & 28.74  & 14.84 & 30.9 \\
   MORE (rand-RA) & 29.33 & 15.54 & 30.73 \\
   MORE & 30.27 & 16.02 & 31.94 \\
    \bottomrule
\end{tabular}
\end{center}
\end{table}

\begin{figure}
\begin{center}
  \centerline{\includegraphics[width=\linewidth]{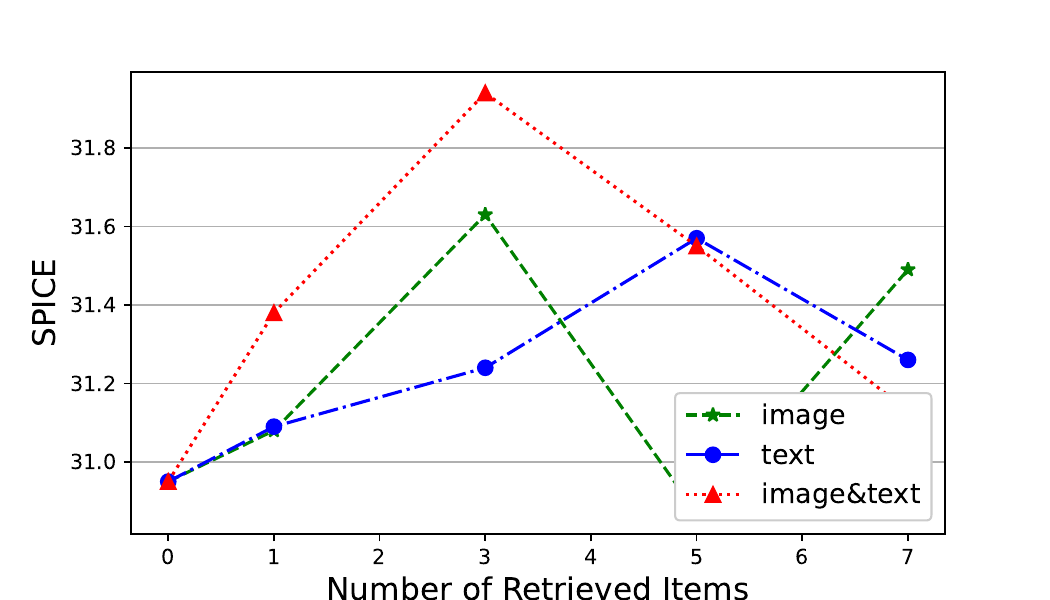}}
  \caption{The SPICE values with respect to the number of retrieved items.}
\label{fig:number}
\end{center}
\end{figure}

\textbf{Are the improvements of MORE attributed to additional parameters?} Considering that our framework introduces more parameters, we investigate whether the performance improvements are attributed to these additional parameters, which arise from two aspects: 1) The retrieval augmented prompt results in an extended input length. To investigate, we adjust the task prompt length from 32 to 64, aligning with the total input length of MORE.
2) The integrator introduces more learnable parameters. To assess whether this would affect performance, we replace the retrieval inputs of each sample with irrelevant retrieved results during training, denoted as rand-RA. This maintains consistency in the learnable parameters with MORE. The experimental results are recorded in Table~\ref{tab:parameter_analysis}. Neither of them shows significant improvements over the backbone T5-base, showing that the benefit of MORE does not come from the extra parameters.

\textbf{Will utilizing more retrieved results enhance the model performance?} In retrieval-augmented methodologies, a crucial factor influencing the final results is the number of retrieval items. We integrate varying numbers of retrieval content for both single-modal and multi-modal settings. The SPICE results are illustrated in Figure~\ref{fig:number}, with additional metric results available in Appendix~\ref{app:other_metric}. No matter which modality is used, the model performance first increases and then decreases as the number of retrieval inputs increases. This is because too few retrieval inputs may lead to insufficient coverage of required information, while an excess of inputs may introduce redundancy and noise.

\begin{figure}
\begin{center}
  \centerline{\includegraphics[width=\linewidth]{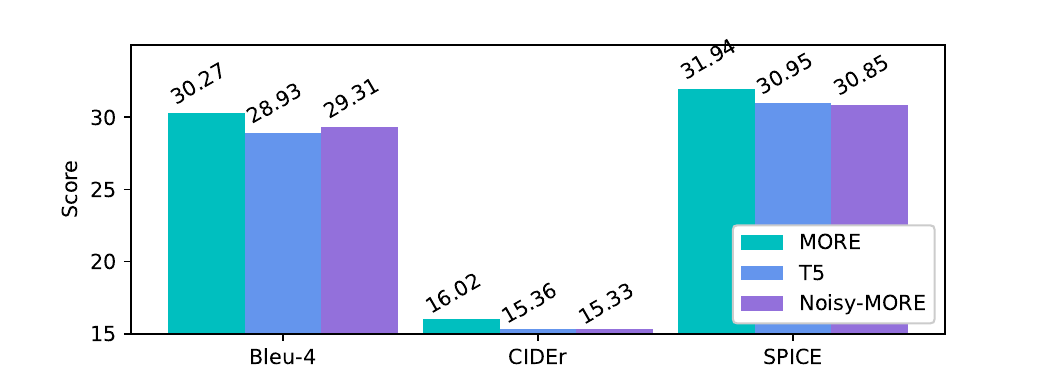}}
  \caption{Test result of the baseline model, MORE augmented with relevant content, and MORE augmented with irrelevant content.}
\label{fig:noisy}
\end{center}
\end{figure}

\textbf{Is MORE robust to noise in retrieval augmentation?} The retrieved results may not always be of high quality and occasionally may be even irrelevant to the query. Therefore it is also important for retrieval augmented models to be robust to the noisy retrieval outcome. We test the model's robustness in the face of poorly retrieved results by feeding it with only irrelevant retrieval content during testing (denoted as Noisy-MORE). As illustrated in Figure~\ref{fig:noisy}, Noisy-MORE performs similarly to its backbone T5 when using irrelevant results for augmentation. This indicates that MORE is robust to the noise in the retrieved items by not blindly trusting the augmentation input.

\subsection{Case Study}

\begin{figure*}
\begin{center}
  \centerline{\includegraphics[width=\linewidth]{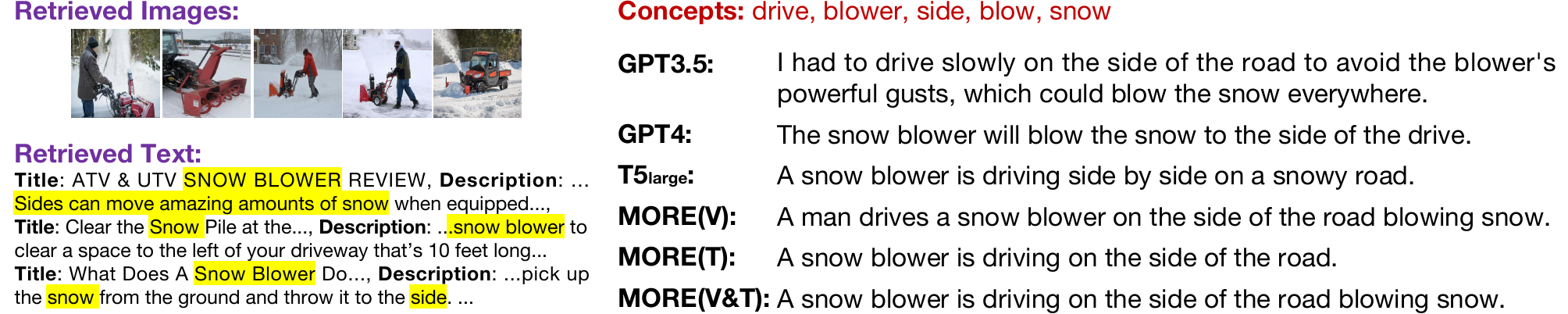}}
  \caption{Generated sentences that benefit from retrieval augmentation}
\label{fig:case_study}
\end{center}
\end{figure*}

We conduct case studies to qualitatively analyze how MORE enhances text generation through retrieval augmentation. As shown in Figure~\ref{fig:case_study}, small LMs like T5$_{\textrm{LARGE}}$ make a nonsensical sentence that `only one blower can not drive side by side'. The retrieved images show a 'blower located on the side of the road' scene, and the retrieved text describes `throw the snow to the side of the road', thereby helping the model clarify the usage of `side' and correct generation errors. More cases can be found in Appendix~\ref{app:case}.

As for LLMs, we find that they sometimes make nonsensical sentences, as shown in Figure~\ref{fig:intro}. This shows that even if a huge amount of parameters and training data are used, the LLMs are still not able to grasp commonsense knowledge completely. Therefore, it is also necessary for LLMs to use retrieval augmentation to provide reference. The other characteristic is that the sentences made by GPT-3.5 and GPT-4 are usually long. To connect the given concepts and output reasonable sentences, they may need more words or information. This also reflects the lack of commonsense knowledge that humans are well aware of. 




\section{Conclusion and Future Work}

To sum up, we introduce MORE, a multi-modal retrieval augmentation framework. Our approach is capable of extracting useful information and disregarding irrelevant noise from visual and textual results of variable quality, thereby assisting language models in generating reasonable sentences. Extensive experiments on the CommonGen task demonstrated the effectiveness of our method. This novel approach may offer a new perspective for retrieval-augmented language models. 

We focus on the generation task in this work and the application of multi-modal retrieval augmentation on more tasks is worth exploring in the future. Besides, the current method concentrates on `how to incorporate multimodal retrieved items' and does not involve optimization of the retrieving step, which is left for future work.

\section{Limitations}

Our method uses soft-prompt,  making it unsuitable for LMs accessible solely through the API as it cannot convey input in natural language form. In addition, to avoid changing the internal structure of the LMs, we adopted the p-tuning in this work. Using more advanced methods such as LoRA~\cite{Hu2021LoRALA} to achieve better results can be considered in future work.

The retrieved results are from public data on the Internet and we did not collect any privately identifiable information in our study. However, it may be inevitable to crawl some public photos and other data, which may, which may still include some personal information, such as faces. We followed Bing's authorization requirements for the use of data and did not modify or use it commercially. We call on anyone using our framework to follow the licensing requirements and not misuse the technology.

\section*{Acknowledgement}
This work was funded by the National Natural Science Foundation of China (NSFC) under Grants No. 62302486, the Innovation Project of ICT CAS under Grants No. E361140, the CAS Special Research Assistant Funding Project, the Lenovo-CAS Joint Lab Youth Scientist Project, the project under Grants No. JCKY2022130C039, and the Strategic Priority Research Program of the CAS under Grants No. XDB0680102.

\bibliography{custom}

\clearpage
\appendix 

\section{Test Result Using Captions}
\label{app:bias_result}
\begin{table}
\caption{Test results on CommonGen(V1.1) by directly using the captions retrieved through BM25 as output and other existing methods.}
\label{tab:bias_result}
\begin{center}
\begin{tabular}{lccc}
    \toprule
    Model & Bleu$_4$ & CIDEr & SPICE \\
    \midrule
    BM25 & 44.07 & 18.64 & 33.47 \\
    DKMR$^2$ & 44.33 & 19.54 & 34.59 \\
    KFCNet & 43.62 & 18.85 & 33.91 \\
    KGR$^4$ & 42.82 & 18.42 & 33.56\\
    \bottomrule
\end{tabular}
\end{center}
\end{table}

Since the Commongen dataset itself relies on caption data during the construction process, and most existing methods use the retrieved caption as a reference for generation, such as DKMR2~\cite{he2022metric}, KFCNet~\cite{Li2021KFCNetKF}, and
KGR4~\cite{liu2022kgr4}. We suspect and test whether the retrieved caption itself reveals the correct answer to some extent. Specifically, we use concepts as query, and then simply use BM25 as the retrieval method to retrieve captions from image captions and video captions. The retrieved caption will be directly used as the prediction result without modification and the results compared with other methods are shown in Table~\ref{tab:bias_result}. It can be seen that even without a tunable retriever and any modification to the caption, good results can be achieved.



\section{Retrieved Inputs Crawling and Preprocessing}
\label{app:crawl}
We concatenate all concepts in a concept set to form a query. For the image, we use the template `a photo of \{...\}' (e.g. a photo of decorate, music, background, and tree) and crawl the first 20 image results returned by the search engine. We further removed duplicate images based on the dHash algorithm. For text, we directly use the concatenation of the concept set as the query. We crawl the text results from the first two pages. Considering that the full document associated with each result may be very long, the search engine has provided a concise text summary of the webpage aligning with the search keywords, we only keep the title and description in the snapshot. We also removed the URL and non-English parts.

\section{Examples of Generated Captions from VisCTG}
\label{app:captions}

\begin{figure}
\begin{center}
  \centerline{\includegraphics[width=\linewidth]{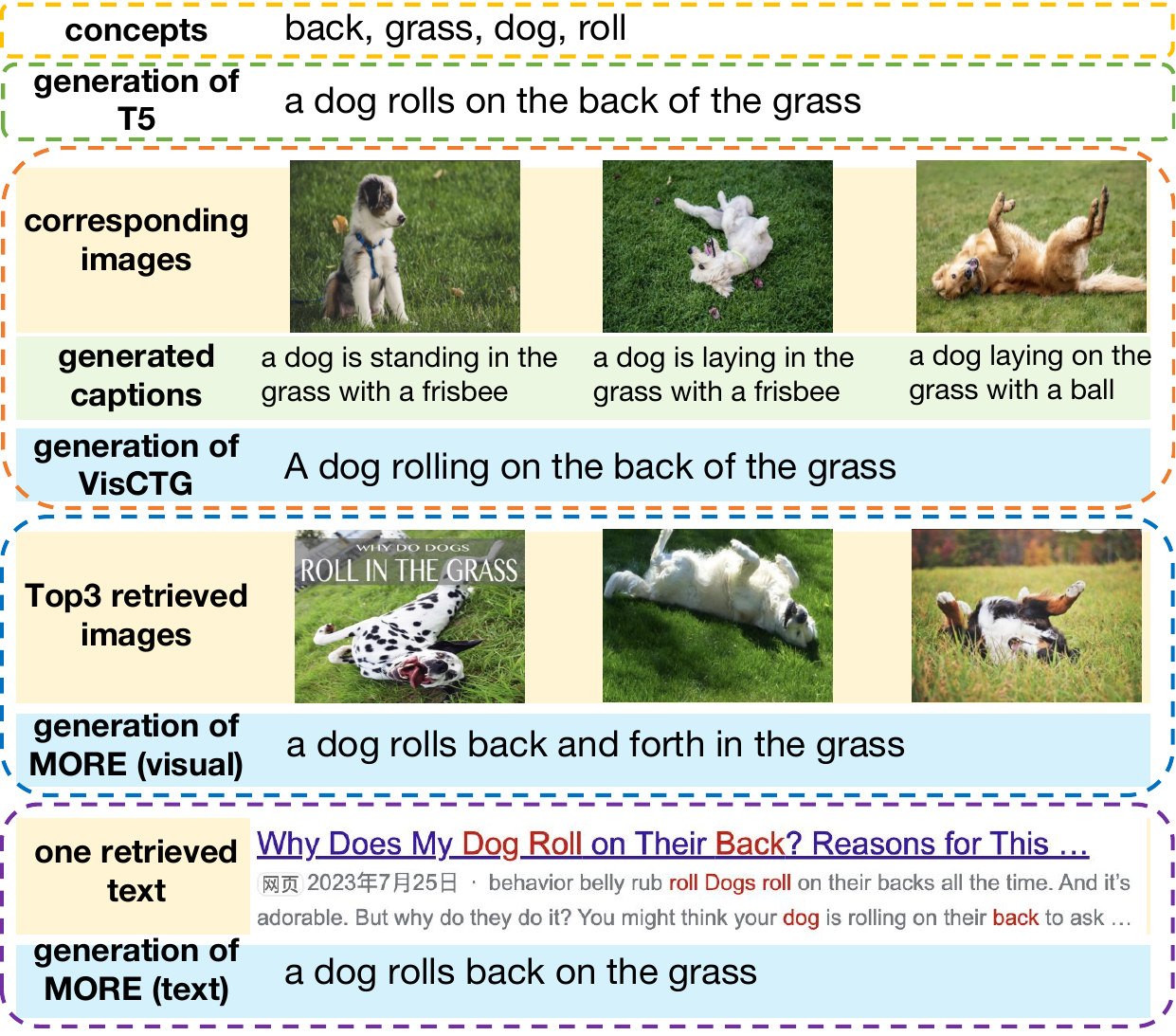}}
  \caption{An example of the generation of VisCTG, the generated captions as well as corresponding images. We also show the generation of MORE and the retrieval content ii use. Since the captions used by VisCTG are ranked by their coverage of the concept words in descending order, the order of images of VisCTG and MORE may be different.}
\label{fig:visctg_example}
\end{center}
\end{figure}

We use an example to illustrate why it is better to directly use the raw image than to convert the image into a caption. There are two main reasons: 1) the generated caption may be inaccurate. As shown in Figure~\ref{fig:visctg_example}, due to the error of the model and the bias in training data, when `dog' appears, the caption model always generates sentences containing `frisbee' or `ball', even though these objects do not appear in the image. Inaccurate captions will further mislead follow-up text generation. 2) the pre-generated captions may lack the required information. In the example, the T5 model incorrectly generates `the back of the grass', and the information needed is `dog rolls on their back`. Although the images contain relevant information, it is not included in the captions, so that the original generation cannot be corrected.

\section{Results of Other Metrics}
\label{app:other_metric}

\begin{figure*}
\begin{center}
  \centerline{\includegraphics[width=\linewidth]{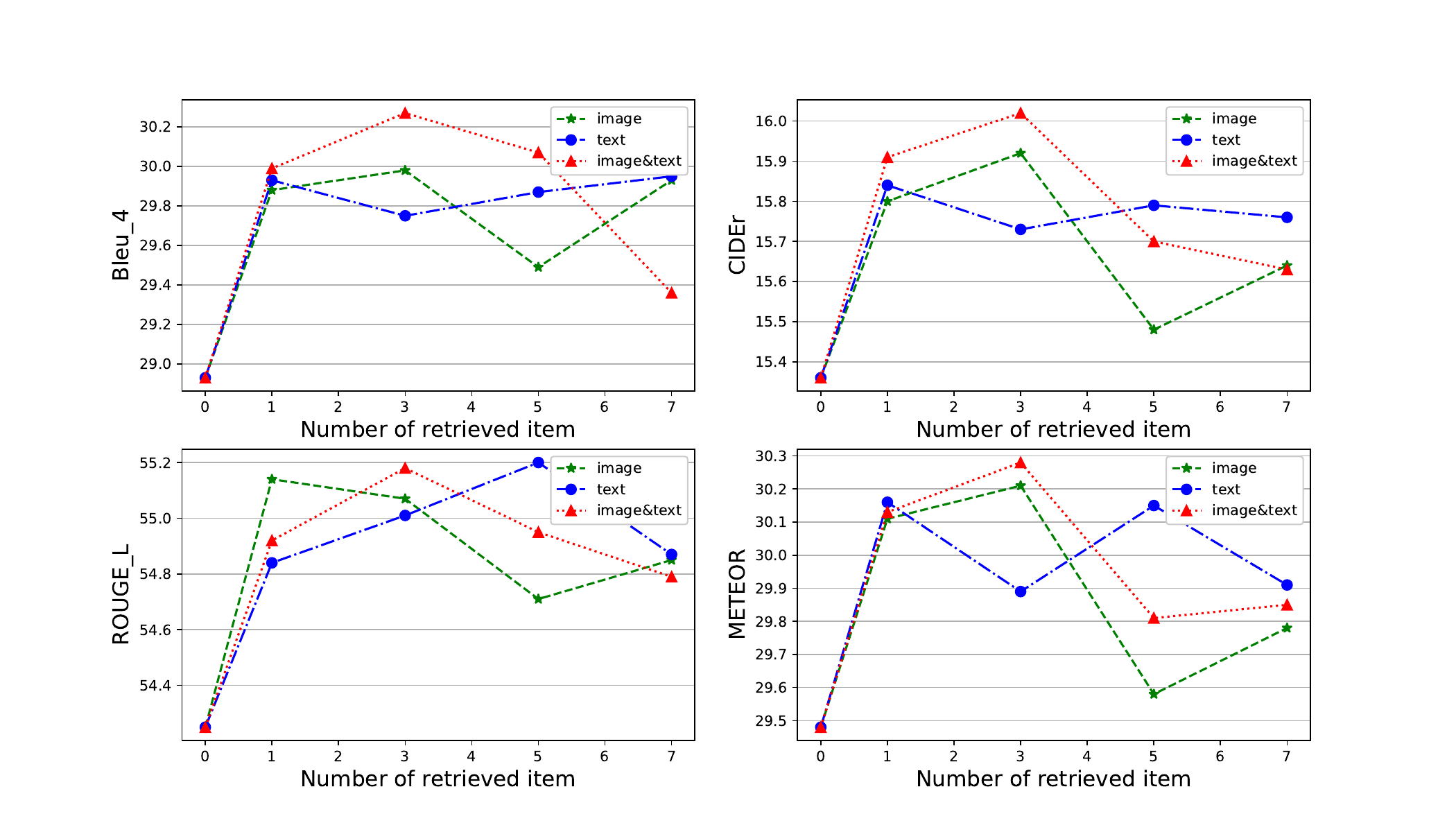}}
  \caption{Scores with different retrieval content numbers.}
\label{fig:other_metric}
\end{center}
\end{figure*}

Combining various metrics shown in Figure~\ref{fig:other_metric}, it can be seen that using three retrieval contents is a better choice. The conclusion that using too many or too few retrieved results will not lead to optimal results has not changed.

\section{Cases of Generation}
\label{app:case}

\begin{figure*}
\begin{center}
  \centerline{\includegraphics[width=\linewidth]{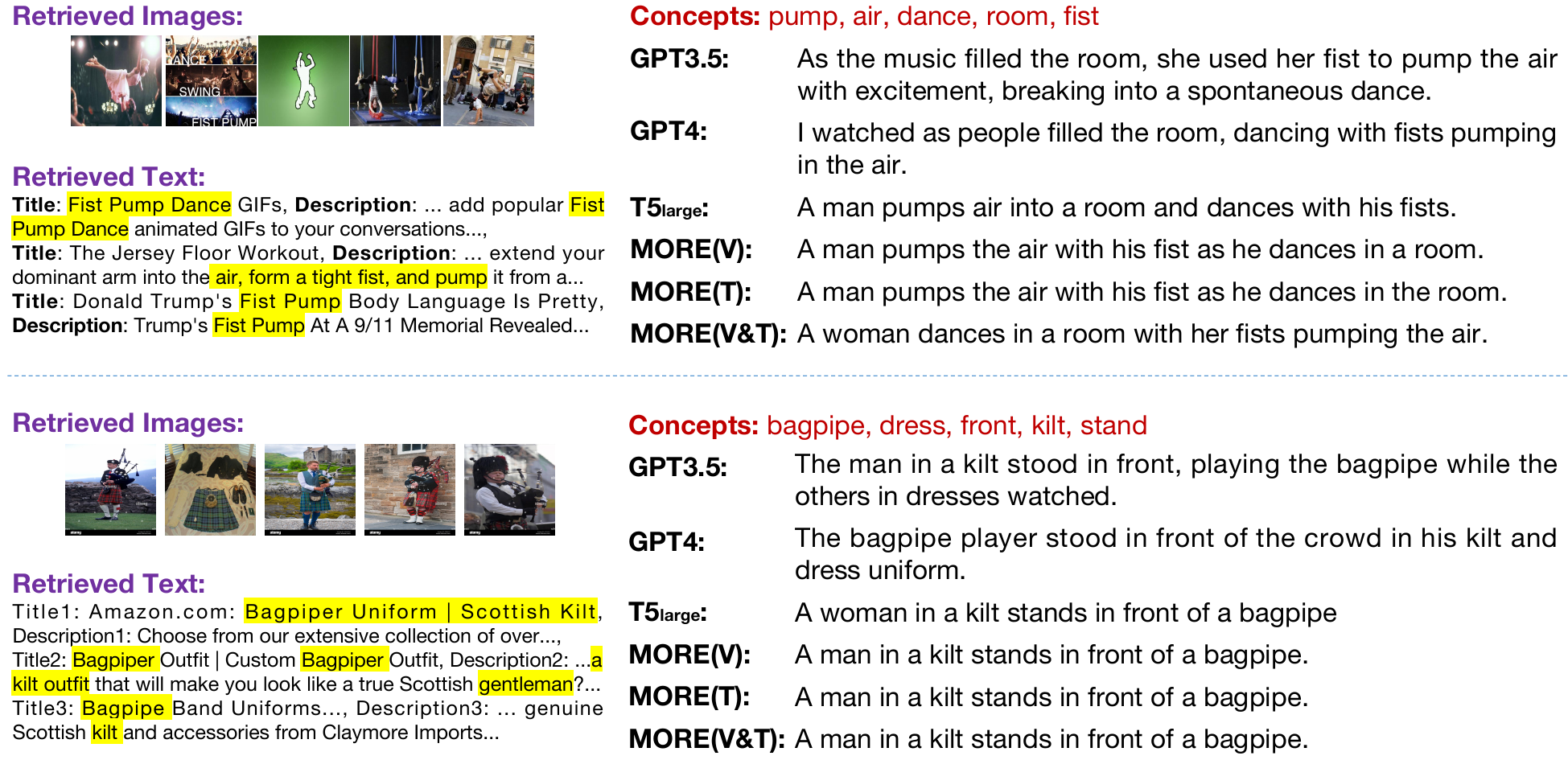}}
  \caption{Generated sentences with/without retrieval augmentation}
\label{fig:2case}
\end{center}
\end{figure*}

We show two additional examples in Figure~\ref{fig:2case} to help intuitively understand how multi-modal retrieval augmentation helps the model generate more reasonable sentences.

\end{document}